\documentclass{article}

\usepackage{microtype}
\usepackage{graphicx}
\usepackage{subcaption}
\usepackage{booktabs} 
\usepackage{float}
\usepackage{natbib}

\usepackage{tikz}
\usetikzlibrary{arrows.meta, positioning, shapes.geometric, fit}

\usepackage{pgfplots}
\pgfplotsset{compat=1.18} 
\usepackage{tikz}
\usepackage{pgfplots}
\pgfplotsset{compat=newest}
\usetikzlibrary{arrows, decorations.pathmorphing, decorations.markings, backgrounds, positioning, fit, petri}
\usepackage{quantikz}

\usepackage{hyperref}

\usepackage{amsmath}
\usepackage{amssymb}
\usepackage{mathtools}
\usepackage{amsthm}
\usepackage{bm}

\usepackage[capitalize,noabbrev]{cleveref}

\theoremstyle{plain}

\theoremstyle{definition}

\theoremstyle{remark}

\usepackage[textsize=tiny]{todonotes}

\begin{document}

 \title{SAFE Quantum Machine Learning with Variational Quantum Classifiers}
%
%
\author{
Ying  Chen, National University of Singapore  \and
Paolo Giudici, University of Pavia, Corresponding author, email: giudici@unipv.it \and
Vasily Kolesnikov, University of Pavia  \and 
Paolo Recchia, National University of Singapore}

\maketitle              %




\begin{abstract}
We propose a variational quantum classifier operating on high-dimensional deep representations via amplitude encoding, stabilized by a learnable classical pre-encoding layer. By combining normalized amplitude embeddings with bounded quantum observables, the resulting model induces a structured and smooth hypothesis class with controlled sensitivity to input variations. Model reliability is assessed using SAFE-AI metrics derived from the Cramér–von Mises divergence, enabling consistent evaluation across accuracy, robustness, and explainability dimensions. Empirical results show that the proposed quantum model provides competitive predictive performance compared with strong classical baselines while exhibiting a more balanced SAFE reliability profile, with improved robustness to noise and stability under structured feature removal. These findings suggest that variational quantum circuits offer a principled mechanism for stability-oriented SAFE learning in safety-critical settings.
 
\end{abstract}

\section{Introduction}

Machine learning systems are increasingly deployed in high-stakes decision-making environments, where failures may have severe societal, economic, or ethical consequences. In such settings, predictive accuracy alone is insufficient: models must also exhibit robustness to input variations and stability under feature or data degradation, while remaining interpretable to human stakeholders. These requirements motivate the growing demand for \emph{Secure, Accurate, Fair, and Explainable} (SAFE) machine learning, in which accuracy, robustness, and explainability are treated as joint design objectives rather than independent post hoc criteria.


While modern deep learning models have achieved remarkable accuracy across a wide range of tasks, their high flexibility often comes at the cost of brittleness under noise, distributional shifts, or the removal of structured features. This tension between expressivity and reliability poses a fundamental challenge for safety-critical applications. As an alternative, we consider hybrid classical--quantum learning models, in which variational quantum circuits naturally operate within constrained, normalized, and bounded hypothesis spaces. A key structural constraint comes from the unitary nature of quantum circuits: each parametrized quantum layer acts as a norm-preserving transformation on the underlying Hilbert space~\cite{nielsen2010quantum,schuld2015introduction,biamonte2017quantum}. In this sense, quantum circuits implement transformations analogous to orthogonal or unitary weight matrices in classical neural networks, where the spectral norm is controlled by construction~\cite{le2015simple,henaff2016recurrent,bansal2018can,li2019orthogonal}. This prevents arbitrary amplification of input perturbations across the quantum layers and imposes an implicit stability constraint on the learned representation.

Through amplitude encoding, classical data are mapped into normalized quantum states, so the input representation is constrained to lie on the unit sphere of the Hilbert space rather than in an unconstrained Euclidean feature space~\cite{schuld2015introduction,schuld2021machine}. The subsequent variational circuit evolves this state through unitary operations, preserving the norm of the quantum state and, before measurement, constraining the transformation to be reversible and inner-product preserving~\cite{nielsen2010quantum}. Finally, expectation-based measurements produce bounded outputs, since observables have finite spectra and expectation values are restricted to a fixed interval~\cite{schuld2021machine,cerezo2021variational}. Therefore, the overall quantum model combines three forms of structural control: normalized input embeddings, unitary feature transformations, and bounded readout functions. These constraints reduce the model's effective freedom and may act as an implicit regularizer, limiting sensitivity to small perturbations and discouraging overfitting.

From this perspective, hybrid quantum models provide a stability-oriented alternative to highly flexible classical architectures. Rather than enforcing orthogonality through explicit re-orthogonalization or spectral penalties, as is often required in classical orthogonal or unitary neural networks~\cite{bansal2018can,lezcano2019cheap,li2019orthogonal}, the quantum component satisfies unitarity by design. This suggests that variational quantum circuits can induce smooth decision functions with controlled sensitivity, offering a natural mechanism for robustness under noise, distributional shift, or structured feature removal. This view is also consistent with recent theoretical studies suggesting that quantum models can exhibit controlled capacity and favorable generalization behavior under appropriate architectural and data-dependent assumptions~\cite{schuld2019quantum,caro2022generalization,abbas2021power, chen2025hybrid}.

In this work, we investigate SAFE quantum learning with variational quantum classifiers and evaluate reliability using a unified family of SAFE-AI metrics derived from the Cramér--von Mises divergence \citep{BabaeiGiudiciRaffinetti2025SRGB}. Using brain tumor classification from MRI images as a representative high-risk example \citep{Menze2015BRATS,Havaei2016Trends}, our empirical analysis shows that variational quantum classifiers achieve predictive performance comparable to strong classical baselines, while exhibiting improved robustness to noise, enhanced stability under data removal, and competitive explainability robustness. These results support the role of quantum models as stability-oriented components for SAFE machine learning systems.

\section{Literature Review}

\subsection{Deep Learning for High-Stakes Analysis}

Deep learning has become the dominant paradigm for high-stakes machine learning tasks due to its ability to learn complex, high-dimensional representations directly from raw data. 

Convolutional neural networks (CNNs) introduced hierarchical feature extraction mechanisms that significantly enhanced the predictive performance across diverse domains, including financial decision-making systems and medical image analysis by learning structured local and global representations from high-dimensional data \cite{Sharma2023OntologyCNN,Huo2024HiFuse}.

Residual architectures further enhanced optimization stability and representational depth, enabling scalable learning in complex settings \cite{mehnatkesh2023intelligent}. More recent self-adapting frameworks, such as nnU-Net, automate architectural design and training procedures while maintaining strong performance guarantees \cite{isensee2024nnu}.

Despite their success, deep neural networks are typically characterized by highly flexible hypothesis spaces and large parameter counts, which can lead to sensitivity under noise, distributional shift, or structured feature removal. These limitations are particularly problematic in safety-critical applications, where model reliability and stability are as important as predictive accuracy. While regularization, data augmentation, and adversarial training have been proposed to improve robustness, such techniques often address isolated failure modes and lack a unified perspective on reliability. This motivates the need for learning frameworks that explicitly balance expressivity with controlled model behavior.

\subsection{Quantum Machine Learning for High-Stakes Analysis}

Quantum machine learning (QML) has emerged as a complementary approach to classical learning models, motivated by the ability of quantum systems to represent and process high-dimensional feature spaces in a compact and parameter-efficient manner. Variational quantum circuits (VQCs), which combine parameterized quantum operations with classical optimization, define structured hypothesis classes through normalized state embeddings and bounded measurement operators \cite{larocca2023theory,gong2024quantum}. Data encoding schemes such as amplitude encoding enable the embedding of exponentially large feature vectors into a logarithmic number of qubits, while expectation-based measurements yield smooth output functions with bounded range \cite{wu2023radio}. 
Recent empirical evidence indicates that hybrid quantum neural networks leveraging amplitude encoding can achieve improved generalization and enhanced stability in high-dimensional, low-sample regimes \cite{chen2025hybrid}, highlighting the benefits of exponential data compression and unitary constraints in noisy learning environments.

Recent studies have further explored hybrid quantum--classical architectures for classification in high-stakes domains. \citep{jabbar2025fusion} proposed a fusion-aware quantum variational autoencoder for biomedical signal classification using amplitude encoding, entanglement-based latent modeling, and hybrid quantum-classical processing. \citep{ajlouni2023medical} proposed a hybrid framework based on classical convolutional methods that uses parameterized quantum circuits for medical imaging tasks.
\cite{felefly2024quantum} conducted a systematic benchmark of quantum-classical models, including variational circuits and quantum annealing approaches, against strong classical baselines. 
\cite{srivastava2025hybrid} introduced hybrid architectures that map classical network parameters into quantum circuits, demonstrating the feasibility of compact quantum models for complex inputs. 

While these works report competitive predictive accuracy, the potential advantages of quantum models in terms of robustness, stability, and controlled sensitivity---key objectives of SAFE learning---remain largely underexplored and are rarely evaluated within a unified reliability framework.
We aim to fill this gap.

\subsection{Deep Learning for Brain Tumor Analysis}

Brain tumor analysis from MRI data has become a widely adopted benchmark for high-stakes learning systems due to its clinical relevance and inherent complexity. Early approaches relied on handcrafted features combined with classical classifiers such as support vector machines  and k-nearest neighbor classifier\citep{kibriya2023novel,basthikodi2024enhancing}. The transition to deep CNN-based methods enabled end-to-end learning and substantially improved performance in tumor detection and segmentation tasks \citep{maani2023advanced,kakarwal2024enhanced,lu2025deep}.
Transfer learning using pretrained architectures such as VGG and ResNet further reduced data requirements while maintaining high accuracy \cite{pillai2023brain}. As a result, brain tumor classification has become a representative testbed for evaluating learning systems not only in terms of accuracy, but also robustness and explainability in safety-critical settings.

\section{Methodology}

\subsection{Quantum Machine Learning Model}

We propose a quantum machine learning framework that integrates deep feature extraction with a variational quantum circuit (VQC) for classification. Through structured quantum feature transformations induced by amplitude encoding and bounded quantum measurements, the framework aims to support SAFE learning objectives by promoting balanced accuracy, robustness, and explainability. The methodology consists of dataset preprocessing, classical feature extraction, quantum feature transformation and classification, and performance evaluation through cross-validated experiments.

A pre-trained ResNet-18 model is utilized as a feature extractor. By the time the network processes an MRI image $I$, it produces a 512-dimensional embedding vector \[ \bm{z} = f_\theta(I), \quad \bm{z} \in \mathbb{R}^{512}, \] where the last fully connected layer of ResNet-18 is replaced by an identity mapping. Feature extraction is performed while the model is in evaluation mode. Unlike dimensionality reduction methods, such as principal component analysis, the proposed QML employs the complete 512-dimensional feature representation.

\paragraph{Pre-VQC Classical Layer.}

Amplitude encoding enables the compact embedding of high-dimensional classical data into a logarithmic number of qubits, but it is known to introduce non-trivial distortions in the data representation~\cite{larose2020robust}. To mitigate these effects, a trainable classical pre-processing layer is introduced prior to quantum encoding. This layer consists of a linear transformation followed by a GELU nonlinearity, allowing the model to learn an adaptive reweighting and reshaping of the feature distribution before amplitude encoding. Such a pre-VQC transformation stabilizes training by aligning the classical feature geometry with the inductive biases of amplitude-encoded quantum states. An ablation study in which this pre-VQC layer was removed resulted in substantially degraded performance and unstable training behavior, highlighting its critical role in the proposed architecture. Similar strategies have been shown to be effective in hybrid quantum–classical architectures, where the inclusion of a learnable classical adaptation stage significantly improves predictive accuracy and robustness under amplitude-encoding constraints~\cite{chen2025hybrid}.

\paragraph{Amplitude Encoding.}
The standardized feature vector $\bm{z} \in \mathbb{R}^{512}$ is normalized 
\[
    \bm{z} \longrightarrow \bm{x} = \frac{\bm{z}}{\|\bm{z}\|},
\]
and then embedded into a quantum state using a 9-qubit amplitude encoding~\cite{schuld2019quantum}, as 
\[
|\psi(x)\rangle = \sum_{i=0}^{511} x_i |i\rangle.
\]
The normalization is enforced to ensure a valid quantum state.
\paragraph{Variational Quantum Circuit Architecture.}
After amplitude encoding, the resulting quantum state is processed by a variational quantum circuit consisting of a single Strongly Entangling Layer acting on 9 qubits. The layer comprises parameterized single-qubit rotation gates, each associated with three variational parameters (the angles $\alpha_i$, $\beta_i$, and $\gamma_i$ defining the rotation $R$ applied to the $i$th qubit), followed by a fully connected pattern of entangling controlled-NOT (CNOT) gates.

The circuit outputs a learned quantum representation by measuring the
expectation values of the Pauli-$Z$ operator on each qubit:
\[
q_i = \langle \psi_{\text{out}} | Z_i | \psi_{\text{out}} \rangle, \quad
q_i \in [-1,1].
\]

\paragraph{Final Linear Layer.}

\begin{figure}[H]
\resizebox{1.\linewidth}{!}{
    \begin{tikzpicture}
        \node[draw, thick, rectangle, fill=red!30] (box1) at (0,0) {
            \begin{tikzpicture}[
            neuron/.style={circle, draw, minimum size=0.8cm}, 
            layer/.style={draw=none,fill=none},
            ->, >=stealth',
            dashedline/.style={draw=black, dashed}
            ]
    
            \node[neuron] (I1) at (0, 3) {};
            \node[neuron] (I2) at (0, 2) {};
            \node[neuron] (I3) at (0, 1) {};
            \node at (0, 0.2) {\vdots};
            \node[neuron] (I4) at (0, -0.7) {};
            \node[layer] at (0, -1.3) {512 ResNet};
        
            \node[neuron] (I21) at (1.5, 3) {};
            \node[neuron] (I22) at (1.5, 2) {};
            \node[neuron] (I23) at (1.5, 1) {};
            \node at (1.5, 0.2) {\vdots};
            \node[neuron] (I24) at (1.5, -0.7) {};
            \node[layer] at (1.5, 3.6) {GELU};
        
            \draw[->] (I1) -- (I21);
            \draw[->] (I1) -- (I22);
            \draw[->] (I1) -- (I23);
            \draw[->] (I1) -- (I24);
            \draw[->] (I2) -- (I21);
            \draw[->] (I2) -- (I22);
            \draw[->] (I2) -- (I23);
            \draw[->] (I2) -- (I24);
            \draw[->] (I3) -- (I21);
            \draw[->] (I3) -- (I22);
            \draw[->] (I3) -- (I23);
            \draw[->] (I3) -- (I24);
            \draw[->] (I4) -- (I21);
            \draw[->] (I4) -- (I22);
            \draw[->] (I4) -- (I23);
            \draw[->] (I4) -- (I24);
            \end{tikzpicture}
        };
        \node[above=5pt of box1] {Pre-VQC};
        
        \node[draw, thick, rectangle, fill=blue!30] (box2) at (7.0, 0) {
            \begin{quantikz}
                \lstick{} & \gate[3]{\ket{\psi}} \gategroup[3,steps=1,style={dashed, inner sep=6pt},label style={label position=above,yshift=0.2cm}]{9 qubits Amplitude} & \qw & \gate{R(\alpha_1, \beta_1, \gamma_1)} \gategroup[3,steps=5,style={inner sep=6pt},label style={label position=above,yshift=0.2cm}]{Strongly Entangling PQC.}   & \ctrl{1} & \qw & \qw & \targ{} & \meter{Z} & \lstick{} \\
                \lstick{} &  & \qw & \gate{R(\alpha_2, \beta_2, \gamma_2)} & \targ{} & \ctrl{1} & \qw & \qw & \meter{Z} & \lstick{}\\
                \lstick{} &  & \qw & \gate{R(\alpha_3, \beta_3, \gamma_3)} & \qw & \targ{} & \qw & \ctrl{-2} & \meter{Z} & \lstick{}
            \end{quantikz}
        };
        \node[above=5pt of box2] {VQC};

        \node[draw, thick, rectangle, fill=red!30] (out) at (13,0)
        {
            \begin{tikzpicture}[
                neuron/.style={circle, draw, minimum size=0.8cm},
                dropout/.style={circle, draw, minimum size=0.8cm, dashed},
                layer/.style={draw=none,fill=none}
            ]
                \node[neuron] (O1) at (0, 0.9) {};
                \node[neuron] (O2) at (0, 0) {};
                \node[neuron] (O3) at (0, -0.9) {};
            \end{tikzpicture}
        };
        \node[above=5pt of out] {3 Classes};
            
        \draw[->] (box1) -- (box2);
        \draw[->] (box2) -- (out);
    \end{tikzpicture}
    }
    \caption{The proposed QML architecture.}
    \label{fig:ampl}
\end{figure}
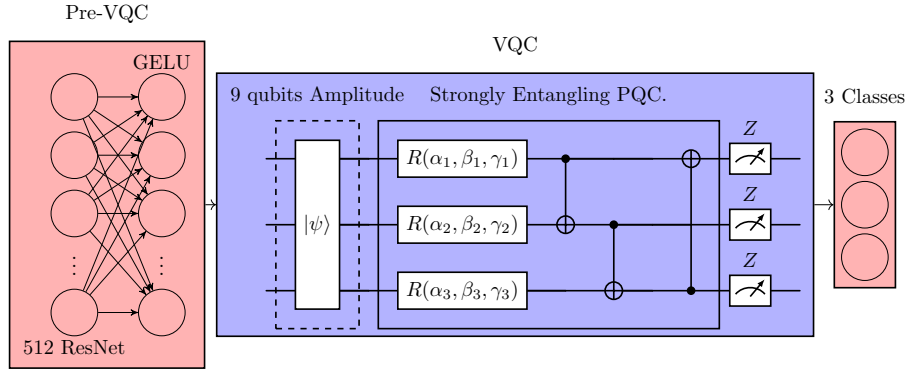

The resulting quantum feature vector $q$ is then mapped
to class logits via a classical linear layer:
\[
y = W_c q + b_c.
\]
Overall, the proposed hybrid quantum–classical model comprises a total of 262k trainable parameters (carried by the pre-VQC linear transformation, followed by the variational parameters in the VQC, and by the final linear layer). The entire hybrid architecture is trained end-to-end using cross-entropy loss, allowing gradients to flow through both classical and quantum components, with the variational quantum circuit evaluated on a noiseless quantum simulator. In Figure~\ref{fig:ampl}, the proposed quantum machine learning pipeline is depicted.

\subsection{SAFE-AI metrics}
To assess the proposed quantum machine learning models and enable a fair comparison with classical baselines, we employ a set of SAFE-AI metrics consistently derived from the Cramér–von Mises divergence.

    Given two random variables, $Y$ and $Y'$, with cumulative distribution functions $F_Y,F_Y':\mathbb{R}\to[0,1]$, we define the $p$-th order Cramér-Von Mises divergence between $F_Y$ and $F_Y'$ as
    \begin{equation}
        CvM_p (F_Y,F_Y') = \int_{-\infty}^{\infty} \big|F_Y(u)-F_Y'(u)\big|^pdF_Y(u)
        \label{Cramér}
          \nonumber
    \end{equation}

 $F_Y$ and $F_Y'$ can be, respectively, the "true" and predicted empirical distribution functions, on a "test" set. But they can also be two predicted distributions, obtained from different data and/or model structures.

We define a Rank Graduation Metric (RG) as a function of the Cramér-Von Mises divergence of order one,  as follows:

    \begin{equation}
      \mathrm{RG}_1 (Y,Y')= 1 - \frac{CvM_1(F_Y,F_Y')}{G(Y)},
            \label{Rga'}
    \end{equation}

    where $G(Y)$ is the Gini index calculated on the true empirical distribution.

    
    When $Y$ is a binary ground truth, and $Y'=\hat{Y}$ the model predictions, the rank graduation metric reduces to the well-known Area Under the ROC Curve $RG=AUC$; see, e.g., \cite{GiudiciRaffinetti2025RGA}. For ordinal outcomes, AUC is no longer applicable, but RG remains well defined and provides a consistent ranking-based measure. When $Y$ is continuous, 
RG can still be computed; however, a more precise reliability metric can be obtained by:
    \begin{equation}
    \mathrm{R^2(Y,Y')} = 1- \frac{MSE(Y,Y')}{Var(Y)}.
\label{R'}    
\end{equation}
where MSE(Y,Y') is the mean squared error between the predicted and true values, leading to the 
the predictive $R^2$.

Note that, while in Equation \eqref{R'} the numerator is the mean squared error of the predictions, in Equation \eqref{Rga'} the numerator is the mean absolute error of the predictive cumulative distributions. While the denominator in Equation \eqref{R'} is the variance, the denominator in Equation \eqref{Rga'} is the Gini Index.  Thus, both Equation \eqref{R'} and \eqref{Rga'} represent the percentage of the variability of $Y$ "explained " by the model; in the former case, the variability is measured by the Variance; in the latter, by the Gini index. 

For multiclass outcomes, the RG metric is computed using a one-vs-rest strategy.
An RG score is evaluated independently for each class by comparing the
corresponding class indicator with the predicted class probability.
The overall multiclass RG is then obtained as a convex combination of the
per-class RG scores, weighted by the empirical class frequencies in the
evaluated split.
 
The $\mathrm{RG}$ metric can be adapted to measure different AI principles, such as robustness, accuracy, and explainability. This can be done by choosing different pairs of input variables, as shown in \cite{BabaeiGiudiciRaffinetti2025SRGB} and summarised below.

\begin{itemize}
    \item \textbf{Accuracy ($\mathrm{RGA}$):} Set $Y$ as the ground truth and $Y'=\hat{Y}$ as the model prediction. 
    Then we denote $\mathrm{RG}$ by $\mathrm{RGA}$, where $A$ stands for accuracy, and use it to measure the overall accuracy of the AI model, $\hat{Y}$.
    \item \textbf{Robustness ($\mathrm{RGR}$):} Set $Y=\hat{Y}$ as the prediction on the original data and $Y'=\hat{Y}^{(p)}$ as the prediction on perturbed data. 
    Then we denote $\mathrm{RG}$ by $\mathrm{RGR}$, where $R$ stands for robustness, and use it to measure the overall robustness of the AI model, $\hat{Y}$. Different choices of perturbation mechanism define different variants of RGR, while preserving the same underlying comparison between original and perturbed predictions.

    \item \textbf{Explainability ($\mathrm{RGE}$):} Set $Y=\hat{Y}$ as the prediction on the full dataset and $Y'=\hat{Y}^{(-j)}$ as the prediction obtained after removing the $j$-th feature(s) (or a group of features).
    Then we denote the $\mathrm{RG}$ by $\mathrm{RGE}$, where $E$ stands for explainability, and use it to assess the explainability of the $j$-th variable(s), with respect to the model
    $\hat{Y}$.

\end{itemize}

 The above derivation underlies the main advantage of the $\mathrm{RGA}$, $\mathrm{RGR}$, $\mathrm{RGE}$  metrics: they are distinct realisations of a common metric family, the $\mathrm{RG}$ metrics, making them consistent with each other, and allowing their integration in a unified assessment metric. The unified metric can be extended to evaluate fairness in terms of the explainability of protected variables, such as gender, race, or nationality. Doing so, and interpreting robustness as security under data perturbations, an overall SAFE metric can be obtained.

\subsection{Implicit Stability and SAFE Learning in Hybrid Quantum Models}


From a modeling perspective, the use of a hybrid quantum machine learning architecture for classification is motivated not only by its expressive power but also by the structural constraints imposed by quantum mechanics on the learned representation. In contrast to fully unconstrained classical neural networks, parametrized quantum circuits operate on normalized quantum states and evolve them through unitary transformations. Therefore, the internal quantum feature representation is constrained to remain on the unit sphere of the Hilbert space throughout the computation. This induces a structured hypothesis class in which the learned transformations are norm-preserving, and the final prediction is obtained through expectation values of bounded self-adjoint measurement operators.

More precisely, amplitude encoding maps a classical input vector into a normalized quantum state. The subsequent variational quantum circuit applies a unitary transformation $U(\phi)$, where $\phi = \alpha_i,\beta_i, \gamma_i$, with $i =1,\dots,9$, indicates the rotation angle parameters in the VQC. The encoded state evolves as
\begin{equation}
\label{eq:quantum_feature_map}
    \ket{\psi(x)} \mapsto U(\phi)\ket{\psi(x)}.
\end{equation}
Since quantum circuits used in the hybrid architecture consist of unitary transformations, they preserve distances at the level of quantum states. For any unitary $U$ and any two quantum states $\ket{\psi}$ and $\ket{\phi}$, one has
\begin{equation}
\label{eq:unit_const}
    \|U\ket{\psi} - U\ket{\phi}\|
    =
    \|\ket{\psi} - \ket{\phi}\|.
\end{equation}
Thus, the quantum feature map is $1$-Lipschitz with respect to its quantum state inputs. In other words, the quantum layer cannot arbitrarily stretch, amplify, or distort differences between internal feature representations. This property is particularly relevant for classification, where small perturbations of the input should not lead to unstable or excessively sensitive changes in the predicted label.

The final output of the quantum model is typically obtained by measuring one or more bounded observables. For an observable $M$, the prediction function can be written in the form
\begin{equation}
\label{eq:q_prediction_function}
    h_\phi(x)
    =
    \bra{\psi(x)} U^\dagger(\phi) M U(\phi) \ket{\psi(x)}.
\end{equation}
If $M$ is a bounded self-adjoint operator, then the output is also bounded. For instance, for Pauli-$Z$ measurements, the corresponding expectation values lie in the interval $[-1,1]$. Therefore, the prediction function is constructed as a composition of normalized state preparation, norm-preserving unitary transformations, and bounded linear functionals. This gives the hybrid quantum classifier an intrinsic form of output control, limiting the extent to which perturbations in the encoded representation can be magnified into unstable predictions.

This structure provides a useful inductive bias. The quantum model is restricted to a hypothesis class that is expressive, but not completely unconstrained. Such a constraint can be interpreted as a form of implicit stability-oriented regularization, analogous to Lipschitz regularization, spectral normalization, or orthogonality constraints in classical deep learning. In particular, classical learning theory has shown that controlling the spectral norms of weight matrices can lead to tighter generalization bounds and reduced effective capacity. For example, Orthogonal Deep Neural Networks impose approximate isometry constraints on the layers, and the most stable case is obtained when the singular values are close to one~\cite{li2019orthogonal}. Similarly, margin-based generalization bounds for deep networks depend on the product of the spectral norms of the weight matrices~\cite{bartlett2017spectrally}. Networks whose layers have spectral norm close to one therefore exhibit stronger control of their Lipschitz constant.

Parametrized quantum circuits naturally satisfy an analogous constraint at the level of the quantum feature representation. Since each quantum layer is unitary, the product of the spectral norms of the quantum transformations is equal to one. As a consequence, the quantum feature extractor is intrinsically spectrally normalized, and the effective complexity of the hybrid model is largely governed by the classical preprocessing and output layers rather than by uncontrolled amplification inside the quantum circuit. This makes hybrid quantum classifiers particularly appealing in settings where one seeks reliable decision boundaries rather than purely high-capacity function approximators.

Unitarity also has implications for optimization stability. During backpropagation, the loss function $\mathcal{L}$ produces a gradient signal with respect to the features produced by the quantum layer. Let
\begin{equation}
\label{eq:loss_gradient_signal}
    g = \nabla_h \mathcal{L}(h_\phi(x))
\end{equation}
denote the gradient signal arriving from the subsequent classical layers. At the level of the quantum state representation, backpropagation through a unitary transformation involves the adjoint unitary $U^\dagger(\phi)$. Since every unitary satisfies
\begin{equation}
\label{eq:unit_g_const}
    \|U^\dagger v\| = \|v\|
\end{equation}
for every vector $v$, the quantum layer cannot amplify the norm of the backpropagated gradient. Hence,
\begin{equation}
\label{eq:quantum_gradient_control}
    \|U^\dagger g\| = \|g\|.
\end{equation}
This provides an intrinsic form of gradient norm control and prevents the quantum feature map itself from causing gradient explosion, a common instability in deep classical networks with unconstrained weight matrices~\cite{hardt2016train}. Although this does not eliminate all potential sources of optimization difficulty in hybrid quantum-classical training, it shows that the quantum component provides a stable, norm-preserving transformation within the overall architecture.

It is important to emphasize that the constraints in Equations~\eqref{eq:unit_const} and~\eqref{eq:quantum_gradient_control} act directly on the internal quantum state representation, not necessarily on the original classical input vector before encoding. The overall Lipschitz behavior of the complete hybrid model also depends on the chosen encoding map, the scaling of the classical features, and the final classical output head. Nevertheless, once the data are embedded into the quantum state space, the parametrized quantum circuit evolves them through strictly norm-preserving transformations and produces bounded measurement outcomes. This creates a meaningful inductive bias: the hybrid quantum classifier is restricted to learn decision functions with controlled sensitivity and bounded responses.

This is especially relevant in SAFE-AI settings, where the goal is not only to maximize predictive accuracy but also to assess robustness, explainability, and reliability. A highly flexible classical model may achieve strong accuracy while remaining brittle under noise, feature perturbations, or distributional shifts. By contrast, the quantum component of the hybrid architecture imposes normalization, boundedness, and unitary evolution by construction. These properties allow the model to be evaluated not merely as a black-box classifier, but as a constrained learning system whose behavior can be characterized in terms of stability and controlled sensitivity. Therefore, proposing a quantum machine learning classifier is justified by the possibility of combining predictive performance with an intrinsically regularized and reliability-oriented hypothesis class.

\section{Application}

We evaluate the proposed hybrid classical–quantum framework on a brain tumor classification task using real MRI images, a representative high-stakes medical application. In this setting, classification errors may lead to delayed diagnosis, inappropriate treatment decisions, or unnecessary clinical interventions, making reliability as critical as predictive accuracy. Beyond correctness, models deployed in such contexts must exhibit robustness to noise and data variability inherent in medical imaging, as well as stable, interpretable behavior to support clinical trust. These requirements directly align with SAFE learning objectives, motivating an evaluation framework that jointly assesses accuracy, robustness, and explainability rather than relying on accuracy alone. The experimental pipeline comprises dataset preparation, deep feature extraction, feature standardization, and cross-validated model evaluation under SAFE-AI metrics.

\subsection{Data and Preprocessing}

Experiments are conducted on the publicly available Brain Cancer – MRI Dataset \cite{Rahman2024BrainCancerMRI}, which contains 6,056 clinically verified MRI images collected from multiple hospitals in Bangladesh. The dataset includes three tumor-related classes: brain glioma (2,004 images), brain meningioma (2,004 images), and brain tumor (2,048 images). All images are provided at a uniform resolution of 
$512\times 512$ pixels and curated in collaboration with medical professionals.

A standardized preprocessing pipeline is applied. First, images are automatically cropped to remove non-informative background and focus on the brain region using contour-based thresholding and morphological operations. Cropped images are resized to $224\times 224$ pixels to match the input requirements of ResNet-18 and normalized channel-wise. High-level representations are then extracted using a pretrained ResNet-18 \cite{He2016ResNet}, yielding a fixed 512-dimensional feature vector for each image. Additional details are provided in Appendix \ref{app:data_preprocessing}.

\subsection{Models and Evaluation Protocol}
All models operate directly on the standardized
512-dimensional feature space, ensuring that differences in performance are attributable to model architecture rather than feature dimensionality. All models operate on this common feature space to ensure fair comparison.

We compare the proposed Quantum Machine Learning (QML) model with several classical models trained on the same features: a linear multinomial logistic regression model with L2 regularization (Linear), a support vector machine with an RBF kernel (SVM), a random forest with 300 trees (RF), and a shallow multilayer perceptron (MLP) with a single hidden layer of 512 units and GELU activation. The MLP architecture mirrors the classical component of the hybrid quantum model (a more detailed description of the models is on~\autoref{app:models}). We underline that the MLP has approximately 264k trainable parameters, slightly higher than in the hybrid QML architecture due to the additional fully connected mapping between the hidden and final output layers. We also emphasize that the MLP architecture was deliberately chosen to match the proposed QML model. In this way, benchmarking the MLP against the QML can be interpreted as an ablation study: the quantum circuit in the hybrid architecture is removed and replaced by a purely classical hidden layer, while the remaining training protocol is kept unchanged. 

All models are evaluated using 5-fold cross-validation. Predictive performance is assessed using macro F1-score and mean squared error (MSE), while reliability is evaluated using SAFE-AI metrics: ranking accuracy (RGA), robustness to noise (RGR), and explainability robustness (RGE), together with their corresponding areas under the curves. We refer the reader to Appendix \ref{app:models} for additional details.

In this work, perturbations are considered at three levels. First, Gaussian noise is added to the standardized ResNet feature vectors to evaluate stochastic feature-space robustness. Second, for differentiable models, we evaluate adversarial feature-space robustness using the Fast Gradient Sign Method (FGSM) \citep{fgsm}. Third, we introduce an image-level spatial robustness test in which MRI images are perturbed through controlled rotations and translations before ResNet feature extraction. This spatial perturbation is gradient-free and model-agnostic with respect to the downstream classifier, since the same perturbed images are processed by all models.

\subsection{Results and Analysis}

\subsubsection{Predictive Performance}

Table~\ref{tab:model_results} reports the predictive performance obtained from 5-fold cross-validation. Among the classical baselines, SVM achieved the highest mean F1-macro score ($0.983 \pm 0.005$), while also exhibiting the lowest mean squared error (MSE) ($0.008 \pm 0.002$), indicating strong and stable probabilistic predictions. The Linear model achieved a mean F1-macro score of $0.938 \pm 0.002$ with a higher MSE ($0.034 \pm 0.002$), whereas RF showed comparable classification performance, with F1-macro $0.934 \pm 0.004$, but the largest MSE ($0.067 \pm 0.001$), reflecting less stable probability estimates.

The proposed QML model achieved a mean F1-macro score of $0.978 \pm 0.004$, closely approaching the performance of the SVM while outperforming the Linear, RF, and MLP baselines. Importantly, QML also attained a low MSE ($0.013 \pm 0.003$), comparable to the strongest classical models and substantially lower than that of the Linear and RF baselines. Overall, these results indicate that the QML architecture achieves competitive predictive performance when compared with other models operating on the same feature representation.

\begin{table}[h]
\centering
\renewcommand{\arraystretch}{1.1}
\setlength{\tabcolsep}{2.8pt}
\begin{tabular}{lcc}
\hline
Model & F1--macro & MSE  \\
\hline
Linear & 0.938 $\pm$ 0.002 & 0.034 $\pm$ 0.002 \\
MLP & 0.959 $\pm$ 0.015 & 0.022 $\pm$ 0.007 \\
RF & 0.934 $\pm$ 0.004 & 0.067 $\pm$ 0.001 \\
SVM & 0.983 $\pm$ 0.005 & 0.008 $\pm$ 0.002 \\
QML & 0.978 $\pm$ 0.004 & 0.013 $\pm$ 0.003 \\
\hline
\end{tabular}
\caption{5-fold cross-validation performance on the MRI dataset. Values are reported as mean $\pm$ standard deviation across folds. Since the dataset is balanced, accuracy and F1-macro yield the same values. Therefore, only F1-macro is reported for compactness.}
\label{tab:model_results}
\end{table}

The observed fold-level variability supports the stability of the QML model. This behavior is expected for high-capacity variational quantum models, which implement expressive non-linear feature maps and entangling operations that effectively lift the data into a much higher-dimensional Hilbert space. Quantum feature maps can create highly expressive hypothesis classes by embedding classical data into exponentially large feature spaces, which naturally increases sensitivity to fold-level variations when the dataset is moderate in size \cite{Schuld2019FeatureMaps}. In the present experiment, however, this sensitivity remains limited, suggesting that the proposed QML architecture maintains stable predictive behavior across folds.

\begin{figure}[!htb]
\centering
\includegraphics[width=\linewidth]{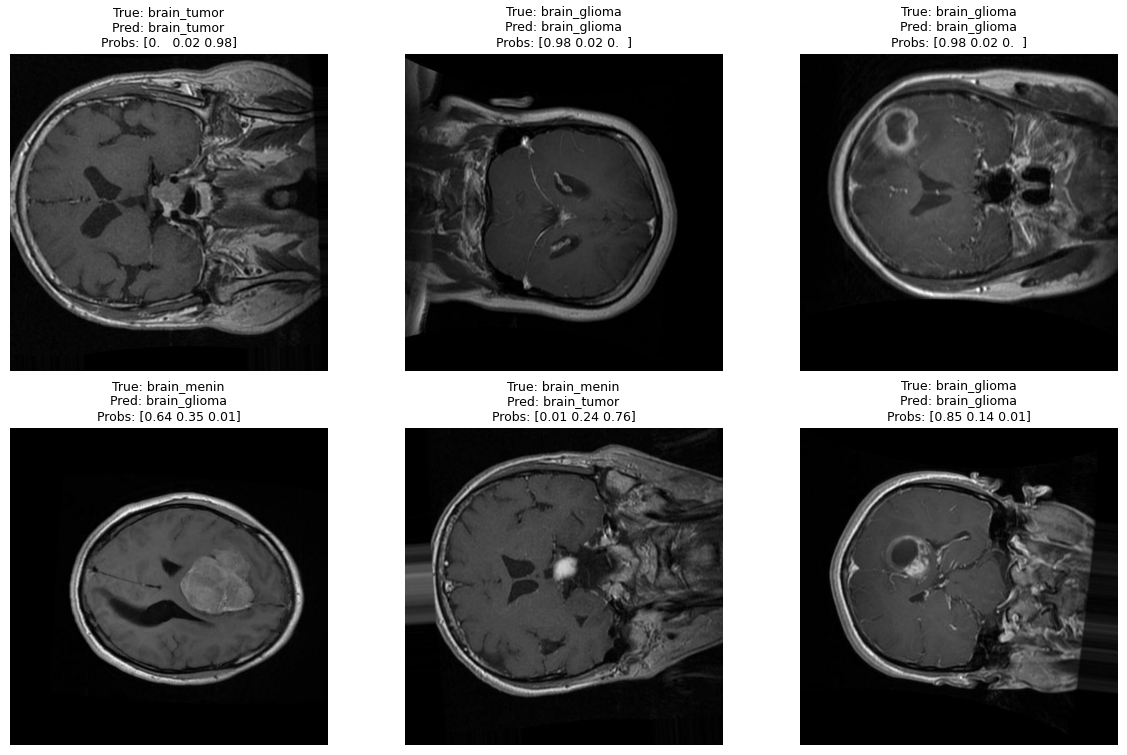}
\caption{Quantum-based tumor classification on unseen MRI scans.
Each image shows the true label, predicted class, and the model’s softmax probabilities}
\label{fig:sample_preds}
\end{figure}

Figure~\ref{fig:sample_preds} presents a qualitative view of the predictions produced by the QML on the given MRI scans from the test set. For each image, the ground-truth label, predicted class, and the corresponding softmax probability vector are reported.
The model assigns a dominant probability to the true class in the correct classification case which indicates that the quantum feature representation captures discriminative tumor characteristics with high confidence score.

The misclassified examples illustrate more challenging scenarios. In particular, some meningiomas are misclassified as gliomas, which can be attributed to overlapping visual characteristics, such as smooth tumor boundaries combined with heterogeneous internal textures. In these cases, the predicted probability mass is more evenly distributed across classes, reflecting increased uncertainty rather than arbitrary misclassification. A small number of examples exhibit high confidence for an incorrect class, which contributes to the observed mean squared error values reported in the quantitative evaluation.

Overall, this qualitative analysis supports the quantitative findings presented before. QML demonstrates strong confidence and consistency in typical tumor presentations, while performance degradation occurs primarily in visually ambiguous or atypical cases. These observations motivate the robustness and explainability analyses conducted using the SAFE-AI framework.

\subsubsection{RGA}
To evaluate the reliability of both models under varying stress conditions, we compute the Area Under Curve (AUC) for each SAFE-AI dimension: accuracy resilience (AURGA), noise-based robustness (AURGR), and explainability stability (AURGE). These values,
reported in Table~\ref{tab:safeai_auc_results}, provide a compact measure of how gracefully model performance degrades under progressive perturbation, allowing a direct comparison of performance.

\begin{figure}[H]
\centering
\includegraphics[width=\linewidth]{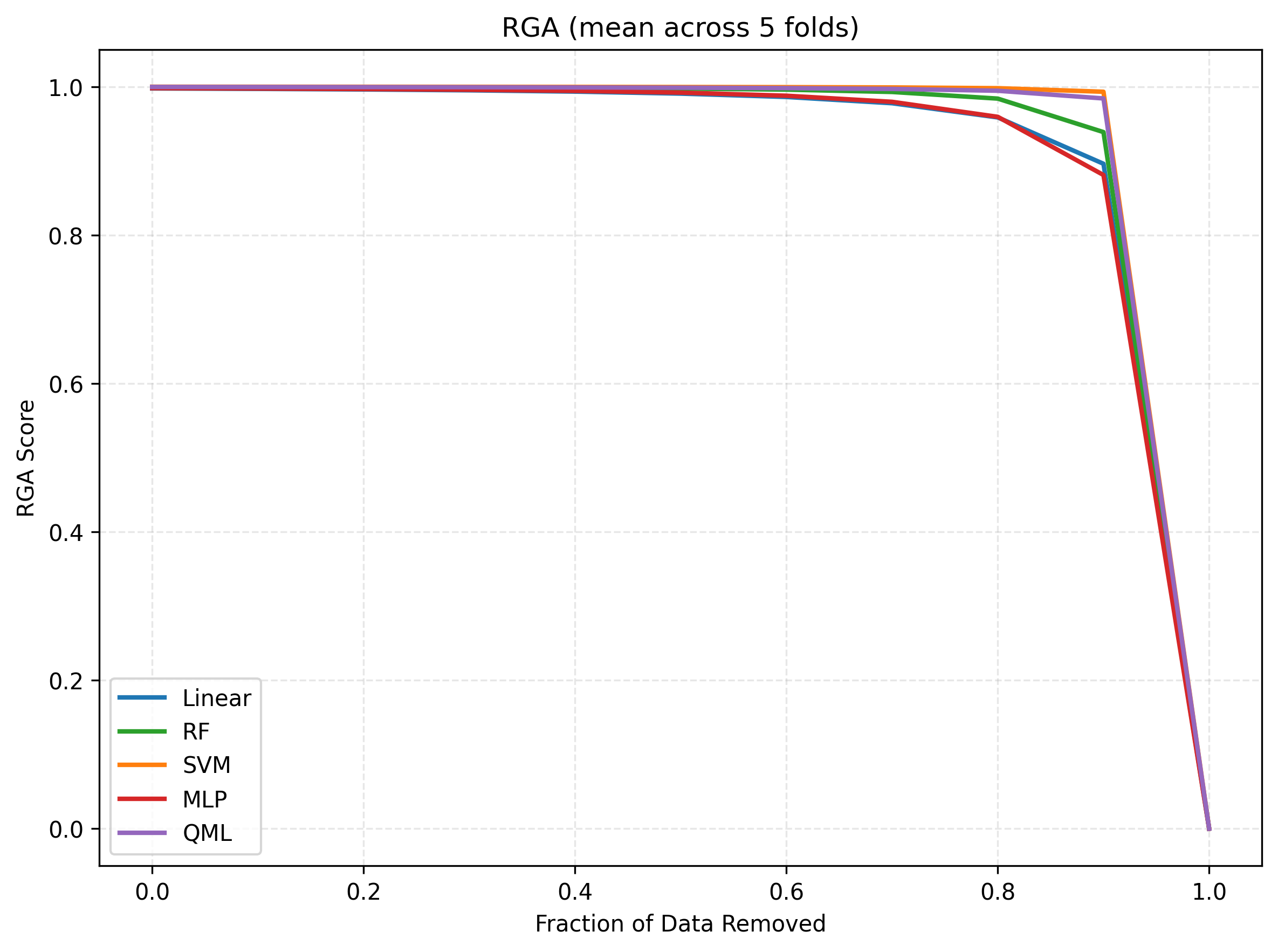}
\caption{\textbf{RGA (mean across 5 folds).}
Degradation of ranking accuracy as a function of the fraction of removed samples.}
\label{fig:rga_mean}
\end{figure}

Figure \ref{fig:rga_mean} shows the RGA curves obtained by removing increasing amounts of data points, ranked according to their predicted confidence. The solid lines represent the average RGA score from 5-fold cross-validation. All models maintain a nearly perfect ranking accuracy when less than 40\% of the data are removed. Beyond this point, the decline varies by model. QML and SVM presented a sharper drop once highly informative samples were taken out, indicating a stronger reliance on top-ranked observations. In contrast, other models experience a more gradual decline, reflecting a more even use of the feature space. The AURGA values support these trends with data, showing a trade-off between the highest ranking accuracy and resilience to significant data removal.

\subsubsection{RGR}
\begin{figure}[H]
\centering
\includegraphics[width=\linewidth]{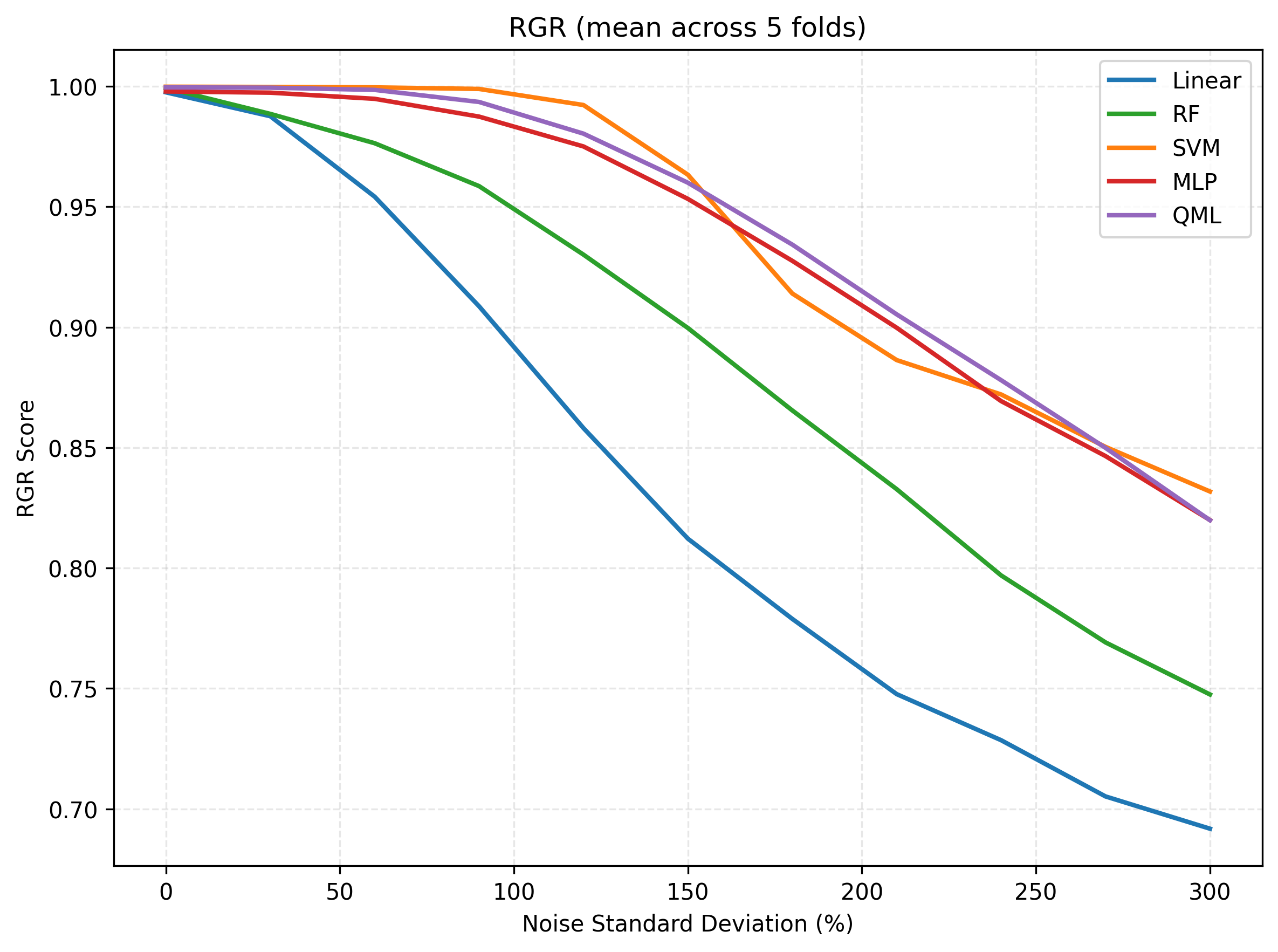}
\caption{\textbf{RGR (mean across 5 folds).}
Robustness to additive Gaussian noise under increasing noise standard deviation.}
\label{fig:rgr_mri}
\end{figure}

Figure \ref{fig:rgr_mri} reports robustness to additive Gaussian noise, measured by the RGR metric. Noise is injected directly into the feature vectors used by the classifiers. The noise level increases from 0\% to 300\% of the signal range, where the signal range is defined as the standard deviation of the test feature values within each cross-validation fold. Solid lines represent mean RGR scores across 5-fold cross-validation.

All models exhibit a monotonic degradation as noise intensity increases, confirming the sensitivity of MRI-based classification to pixel-level perturbations. However, clear differences emerge across learning paradigms. The QML achieves the highest overall robustness, as reflected by the largest AURGR value, indicating a slower and more controlled degradation under increasing noise. This suggests that the quantum feature transformation yields smoother and more noise-tolerant decision boundaries.

The MLP and Support Vector Machine display similar robustness profiles at low-to-moderate noise levels but degrade more rapidly under severe perturbations. Random Forests exhibit an intermediate behavior, while the Linear model shows the fastest decline, highlighting its sensitivity to stochastic input variations.

These findings are consistent with SAFE-AI principles, where models with compact and regularized representations may sacrifice peak accuracy while gaining improved robustness to random perturbations. The RGR analysis therefore highlights the QML as the most noise-resilient model for the MRI classification task.

To further investigate robustness beyond Gaussian feature noise, we performed an additional analysis using FGSM feature-space perturbations and image-level spatial perturbations. FGSM is implemented through the IBM Adversarial Robustness Toolbox (ART) \citep{art}. In our main comparison, FGSM is reported for the Linear model, MLP, and QML, for which input-gradient-based feature perturbations are directly supported in the implemented evaluation setting. The attack is applied to the standardized ResNet feature vectors, and the perturbed predictions are compared with the original predictions using the RGR metric.

In addition, we evaluate spatial robustness at the image level. For each fold, a stratified subset of MRI images is perturbed before feature extraction. Spatial perturbations are generated by applying controlled rotations and translations with increasing perturbation strength. For each strength level, several rotations and translations are considered, and the resulting perturbed images are passed through the same ResNet-18 feature extractor and standardization pipeline used for the original images. The downstream classifiers are then evaluated on the perturbed feature representations. This procedure is gradient-free and model-agnostic with respect to the downstream classifier, since the same transformed images and extracted features are used for all models.

The results, reported in Table~\ref{tab:advers} and Figure~\ref{fig:rgr_mri}, provide complementary views of robustness. The FGSM experiment evaluates local adversarial sensitivity in the standardized feature space, while the spatial experiment evaluates the stability of the full image-to-feature pipeline under controlled geometric transformations. In FGSM, the attack level corresponds to the feature-space perturbation budget $\epsilon$, whereas in the spatial experiment it controls the maximum rotation and translation strength applied to the original MRI images before feature extraction. These perturbation scales are therefore not directly comparable, but are chosen to produce interpretable robustness curves within each setting. Under FGSM perturbations, QML achieves the highest robustness among the evaluated models. Under spatial perturbations, all models remain relatively stable across moderate transformation strengths, with SVM and MLP achieving the highest AURGR-Spatial values.

\begin{table}[!htb]
\centering
\begin{tabular}{lcc}
\hline
\textbf{Model} & \textbf{AURGR-FGSM} & \textbf{AURGR-Spatial} \\
\hline
Linear & 0.4508 $\pm$ 0.0057 & 0.8938 $\pm$ 0.0095 \\
RF     & --                  & 0.9240 $\pm$ 0.0130 \\
SVM    & --                  & 0.9389 $\pm$ 0.0130 \\
MLP    & 0.6683 $\pm$ 0.0224 & 0.9248 $\pm$ 0.0091 \\
QML    & 0.6961 $\pm$ 0.0062 & 0.9163 $\pm$ 0.0163 \\
\hline
\end{tabular}
\caption{Supplementary robustness analysis using FGSM feature-space perturbations and image-level spatial perturbations. Values are reported as mean $\pm$ standard deviation across folds.}
\label{tab:advers}
\end{table}

\begin{figure}[H]
\centering
\includegraphics[width=0.85\linewidth]{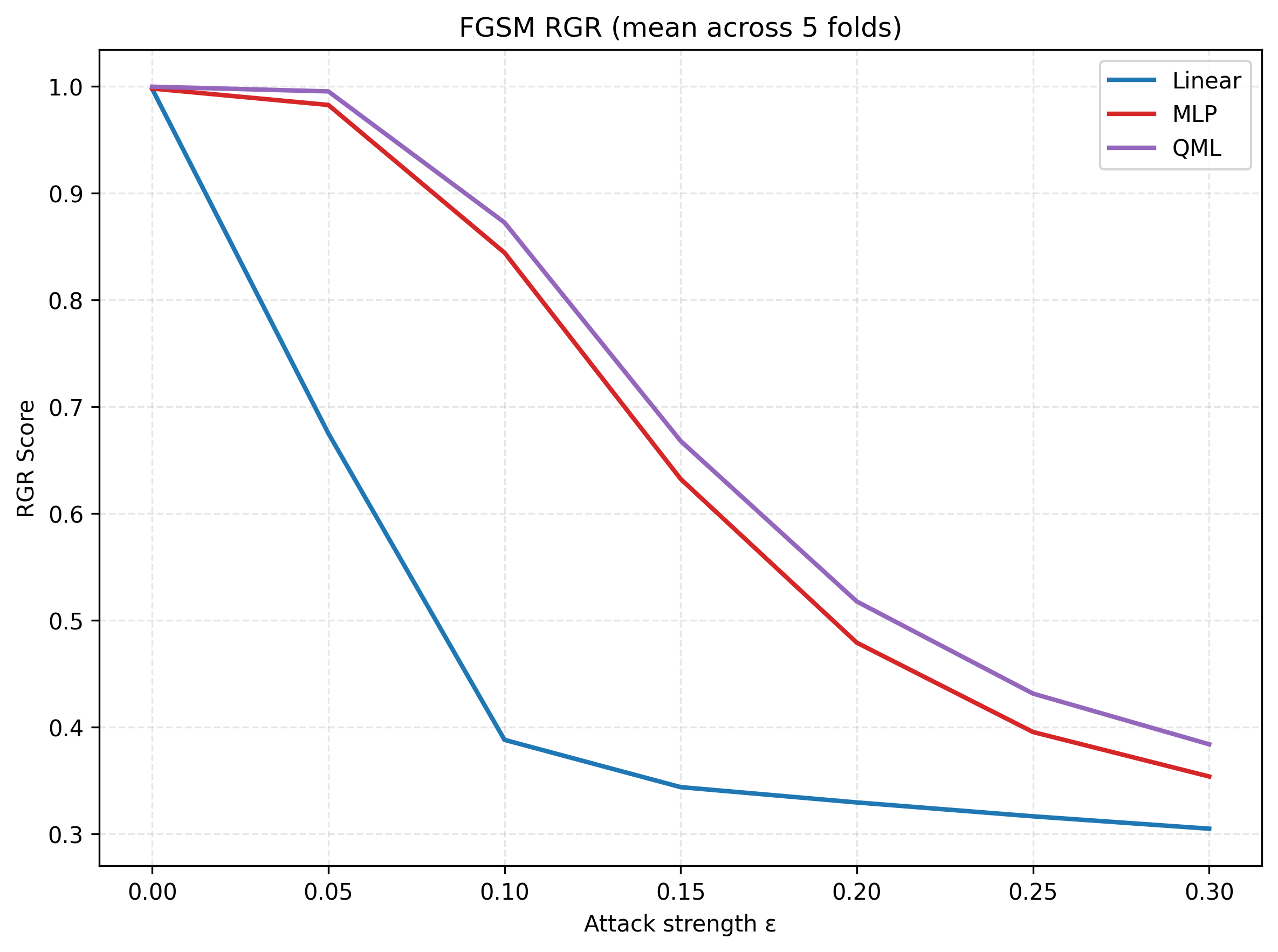}

\vspace{4pt}

\includegraphics[width=0.85\linewidth]{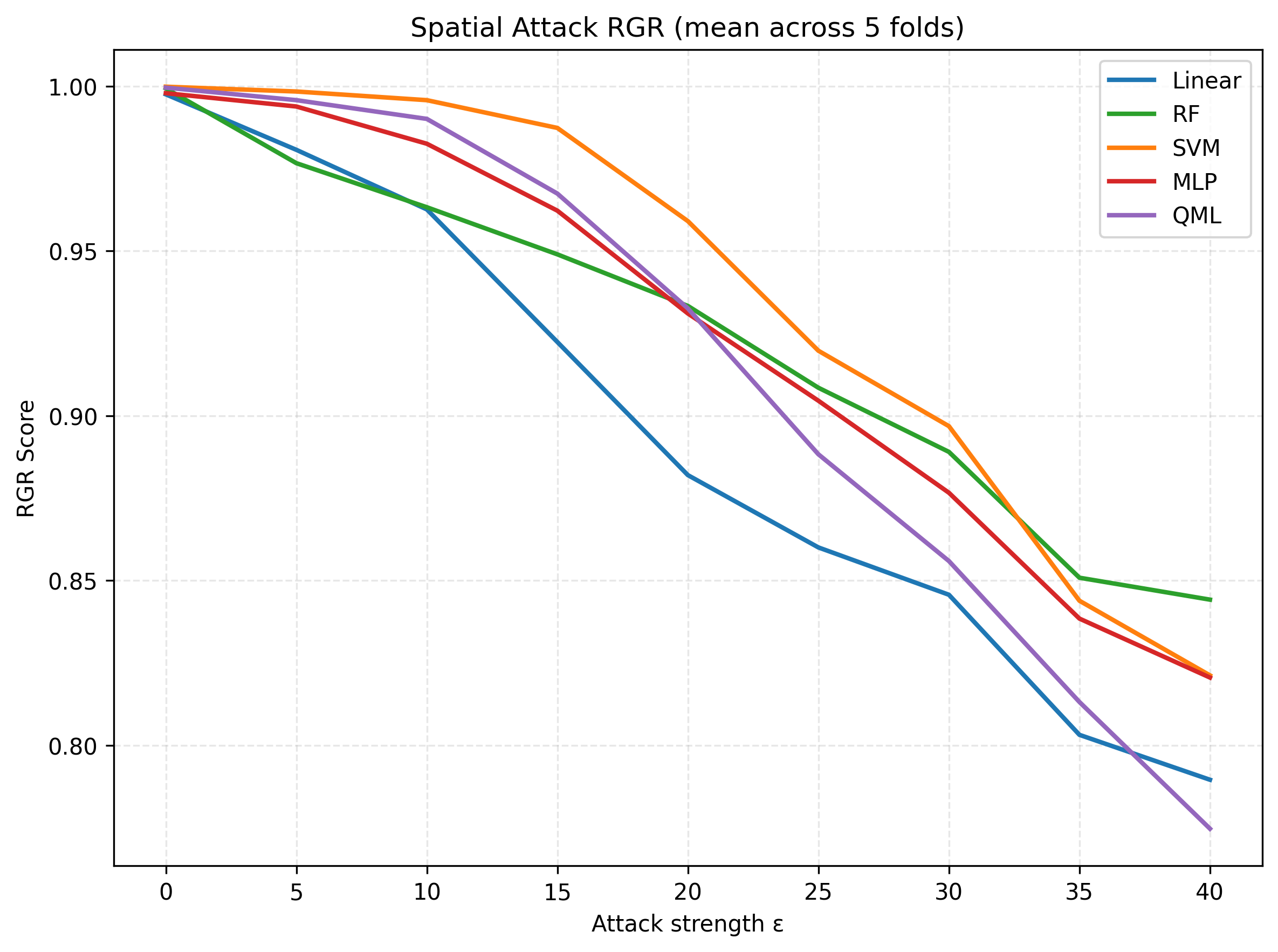}
\caption{Robustness curves under FGSM feature perturbations (top) and spatial image perturbations (bottom)}
\label{fig:supplementary_rgr}
\end{figure}

\subsubsection{RGE}
For the RGE analysis on MRI data, explainability is evaluated directly at the input image level through a structured occlusion process. The proposed approach operates on image pixels, ensuring that explainability is assessed with respect to meaningful and visualizable input degradation.

To guide the occlusion process, we first compute spatial importance maps using a Grad-CAM probe model built on top of the same feature extractor (ResNet-18) used across all classifiers. This probe consists of a frozen feature extractor followed by a linear classification head trained on the true labels. The resulting Grad-CAM heatmaps provide an image-level ordering of spatial regions according to their relevance.

Occluded patches are replaced with a Gaussian-blurred baseline, obtained using
a separable Gaussian kernel (kernel size $31$, $\sigma$ = 7.0) applied
independently to each channel.

\begin{figure}[!htb]
\centering
\includegraphics[width=\linewidth]{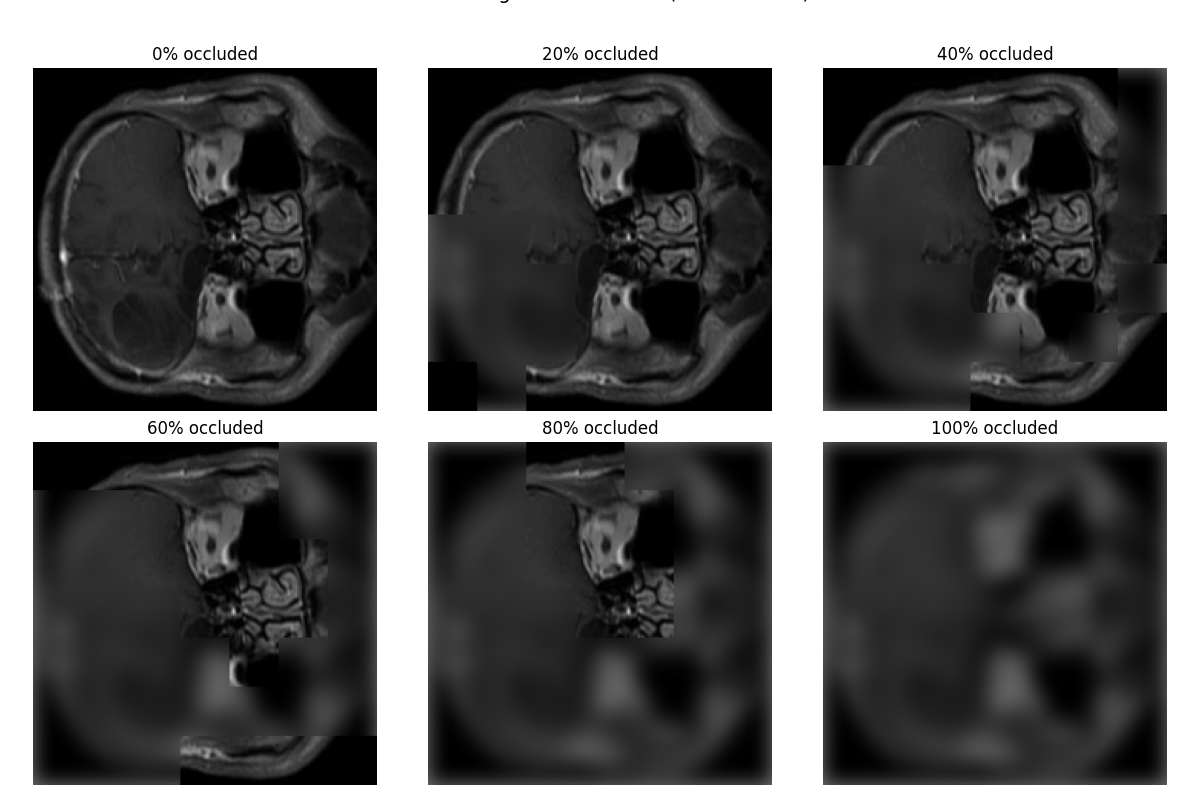}
\caption{\textbf{Example of progressive image occlusion used in the RGE analysis.}
The original MRI image is shown alongside increasingly occluded versions obtained by masking the most relevant regions according to Grad-CAM–based patch ranking.}
\label{fig:rge_occlusion_example}
\end{figure}

Given an input image of size $H \times W$, the image is partitioned into non-overlapping square patches of size $p \times p$. For each image, patches are ranked according to the average Grad-CAM relevance within the patch (see e.g. \cite{Selvaraju2017GradCAM}. For a given occlusion level $k \in [0,1]$, the number of patches to be removed is defined as
\begin{equation}
K = \left\lfloor \frac{k \cdot H \cdot W}{p^2} \right\rfloor,
\end{equation}
corresponding to an expected occluded image area of approximately $k$ times the total image area. The top-$K$ most relevant patches are then progressively removed by blurring, yielding a controlled and monotonic degradation of the input. Figure~\ref{fig:rge_occlusion_example} shows an example of an input image under progressive occlusion.

For each occlusion level, the class probability predictions obtained from the occluded images are compared to those from the original images using the RGE metric. Although the SAFE framework refers to the removal of features in a general sense, in the imaging context, this corresponds to the removal of spatially localized input information, allowing RGE to quantify the stability of model predictions under progressive degradation of the most relevant image regions.

Importantly, the relevance ordering is shared across all models, ensuring that the occlusion strategy is model-agnostic, while the explainability evaluation itself remains model-specific.

\begin{figure}[H]
\centering
\includegraphics[width=0.9\linewidth]{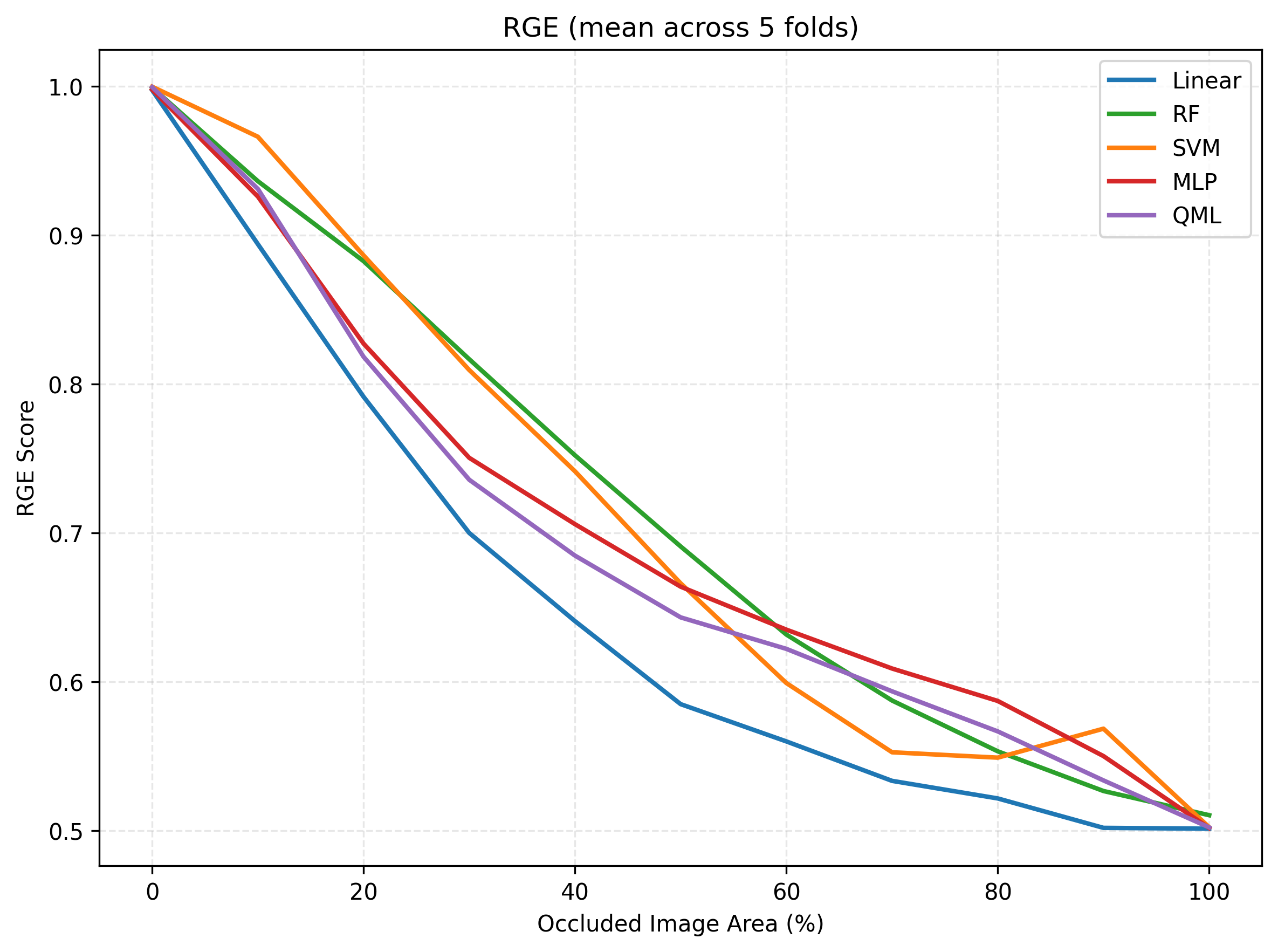}
\caption{RGE explainability robustness curves on the MRI dataset under progressive occlusion.}
\label{fig:rge_curve}
\end{figure}

Figure \ref{fig:rge_curve} illustrates explainability robustness under progressive patch-based image occlusion. Random Forest achieves the highest overall RGE, indicating strong stability of predictions when visual information is systematically removed. The QML improves upon the Linear baseline and remains competitive, particularly at moderate occlusion levels. The Support Vector Machine exhibits a similar degradation trend, while the Linear baseline shows the fastest decline, reflecting greater sensitivity to feature removal.

\subsubsection{Overall SAFE-AI Comparison}

\begin{table}[!htb]
\centering
\renewcommand{\arraystretch}{1.1}
\setlength{\tabcolsep}{3.2pt}

\resizebox{\linewidth}{!}{%
\begin{tabular}{lcccc}
\hline
\textbf{Model} & \textbf{RGA} & \textbf{AURGA} & \textbf{AURGR} & \textbf{AURGE} \\
\hline
Linear & 0.9976 $\pm$ 0.0003 & 0.9289 $\pm$ 0.0022 & 0.8326 $\pm$ 0.0027 & 0.6477 $\pm$ 0.0171 \\
RF & 0.9994 $\pm$ 0.0001 & 0.9404 $\pm$ 0.0017 & 0.8891 $\pm$ 0.0006 & 0.7132 $\pm$ 0.0048 \\
SVM & 0.9999 $\pm$ 0.0001 & 0.9488 $\pm$ 0.0006 & 0.9393 $\pm$ 0.0052 & 0.7090 $\pm$ 0.0074 \\
MLP & 0.9979 $\pm$ 0.0017 & 0.9283 $\pm$ 0.0182 & 0.9360 $\pm$ 0.0056 & 0.7004 $\pm$ 0.0163 \\
QML & 0.9996 $\pm$ 0.0002 & 0.9469 $\pm$ 0.0014 & 0.9409 $\pm$ 0.0013 & 0.6880 $\pm$ 0.0033 \\
\hline
\end{tabular}
}
\caption{SAFE-AI metric comparison across models on the MRI dataset. Values are reported as mean $\pm$ standard deviation across folds}
\label{tab:safeai_auc_results}
\end{table}

To quantitatively support these visual trends, Table~\ref{tab:safeai_auc_results} reports the SAFE-AI metrics across all considered dimensions, including ranking accuracy (RGA) and the corresponding areas under the robustness and explainability curves. The results highlight complementary behaviors across models. The SVM achieves the highest ranking accuracy (RGA) and the strongest robustness to data removal (AURGA), the QML achieves the highest robustness to Gaussian input perturbations (AURGR), and the Random Forest provides the strongest explainability stability (AURGE). 
These differences illustrate the complementary strengths of classical and quantum models across the SAFE-AI dimensions.

\begin{figure}[!htb]
\centering
\includegraphics[width=0.82\linewidth]{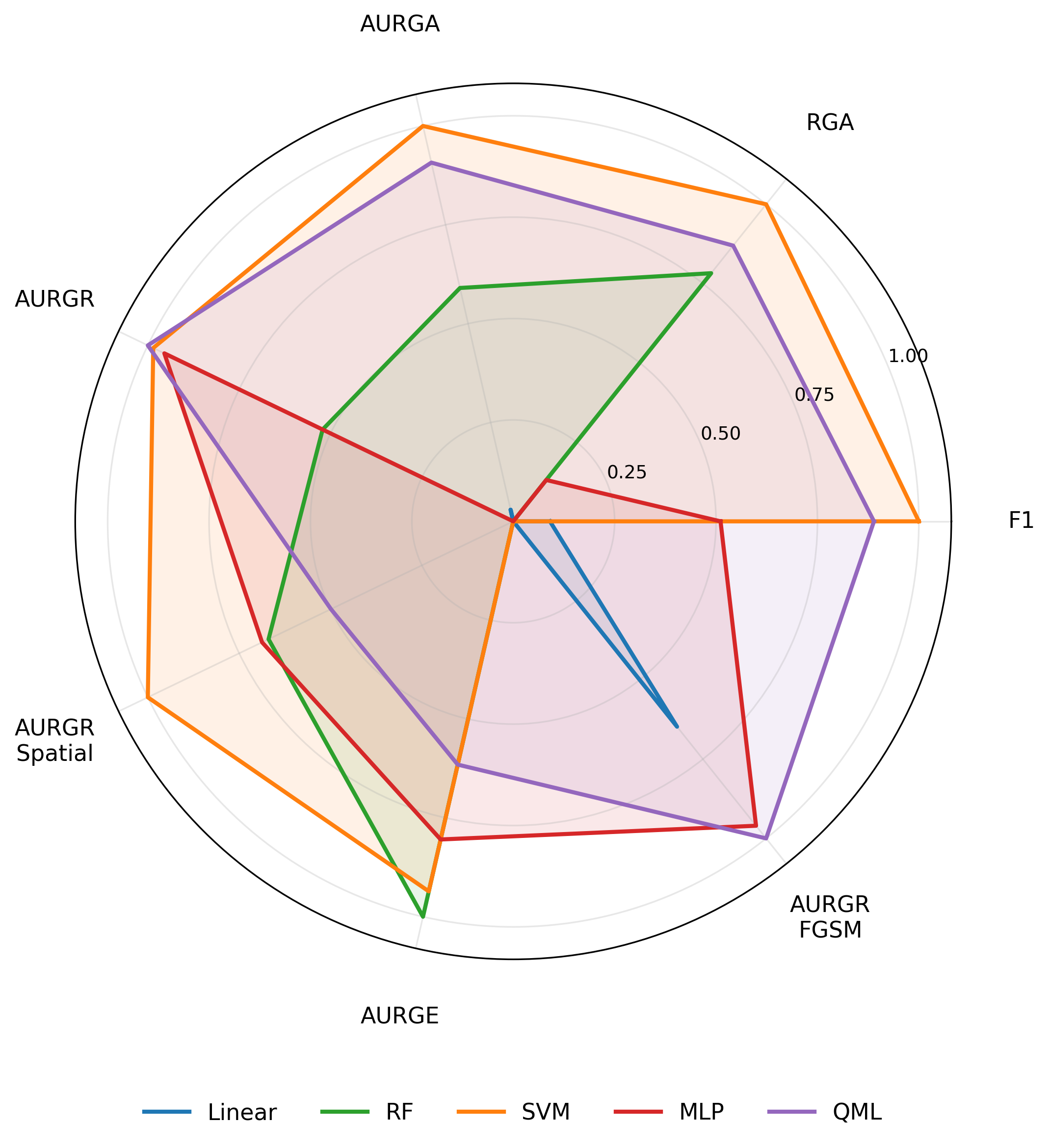}
\caption{Radar plot summarising F1-macro and SAFE-AI metrics. All metrics are min--max normalized across models. FGSM robustness is evaluated only for the Linear, MLP, and QML models. For visualization, RF and SVM are assigned zero on the FGSM axis}
\label{fig:radar}
\end{figure}

Figure~\ref{fig:radar} summarizes the complementary behavior of the models across predictive and SAFE-AI dimensions in a radar plot. The SVM achieves the strongest overall profile in terms of F1-macro score, ranking stability, and spatial robustness. Random Forest provides the highest explainability stability, as reflected by AURGE. The QML remains highly competitive, achieving the strongest robustness to Gaussian feature noise and the highest FGSM robustness among the models for which this attack is evaluated, while also maintaining high F1-macro performance and ranking stability.

Despite the close architectural resemblance between the MLP and the proposed QML, both operating on the same 512-dimensional feature space and with the MLP having more trainable parameters, the QML consistently demonstrates stronger and more balanced performance across almost all considered evaluation metrics, except AURGE and spatial AURGR. This suggests that the inclusion of the variational quantum layer may provide a useful stability-oriented inductive bias compared to replacing it by an additional purely classical nonlinear mapping.

The comparison also shows that the QML advantage is more pronounced for perturbations applied directly to the extracted feature representation, such as Gaussian noise and FGSM, than for perturbations applied to the original images before feature extraction. In particular, AURGR-Spatial and AURGE operate at the image level before ResNet feature extraction, and these tests affect the QML relatively more strongly.
        
\section{Conclusions}

This paper introduces a hybrid classical–quantum machine learning framework for high-dimensional image classification, combining deep feature extraction via ResNet-18 with a Quantum Machine Learning Model (QML) comprising a classical preprocessing layer and a Variational Quantum Classifier (VQC). By embedding the standardized 512-dimensional feature representations directly into the proposed QML pipeline, the approach avoids explicit dimensionality reduction, thereby effectively preserving the full set of features extracted by ResNet-18.

To assess model performance beyond standard predictive metrics, we evaluated both quantum and classical models using SAFE-AI metrics, which jointly quantify accuracy, robustness, and explainability. The empirical results on a brain tumor MRI classification task reveal a nuanced performance landscape with complementary strengths across models. In particular, the SVM achieves the strongest overall predictive performance and ranking stability, while Random Forests provide the highest robustness to explainability. The proposed QML attains F1-macro performance close to the best classical baseline and exhibits the strongest robustness to Gaussian perturbations, together with the highest adversarial robustness among differentiable models. These results highlight its role as a stability-oriented alternative within SAFE-AI evaluation rather than as an approach aimed solely at improving predictive performance over classical methods.

Qualitative analysis further indicates that the QML produces confident predictions for typical tumor presentations, with uncertainty primarily arising in visually ambiguous cases. These findings suggest that the value of the proposed quantum model lies not in surpassing classical methods in peak accuracy, but in providing a compact, structured hypothesis class that supports SAFE learning objectives in high-stakes settings under both stochastic and adversarial perturbations.

\section{Impact statement}

This work demonstrates to the machine learning community the promise of hybrid quantum–classical models as reliability-aware components in image analysis pipelines.

We deliberately do not report computational time comparisons across models. All classical baselines and the QML architecture were trained and evaluated on the same classical hardware, with the quantum component executed on a noiseless quantum simulator. Under this setting, any wall-clock benchmarking would be inherently biased toward classical models and would not reflect the computational characteristics of quantum execution on actual quantum hardware.

Future work will extend this framework to hardware-efficient quantum circuits, multimodal MRI inputs, and larger clinical datasets, with the aim of further investigating the practical viability and deployment potential of quantum-enhanced SAFE learning systems. In this context, an important direction will be the assessment and mitigation of errors arising from execution on noisy quantum hardware, including the impact of gate noise, finite sampling, and device-specific imperfections on both predictive performance and SAFE-AI robustness metrics.

\section*{Software and Data}

All code used for data collection, preprocessing, model training, and evaluation in this study is publicly available and can be accessed at:
\url{https://github.com/koleso500/safe-qml}.
This repository contains scripts and instructions to reproduce the experimental results reported in this paper.

\bibliography{example_paper}
\bibliographystyle{icml2026}

\appendix
\section{Appendix: Additional Experimental Details}

\subsection{Data and Pre-processing}
\label{app:data_preprocessing}




All MRI images are preprocessed using a standardized pipeline. First, an automatic cropping procedure is applied to remove non-informative background regions and focus on the brain area. Cropping is a widely adopted preprocessing strategy in studies involving raw MRI scans, as it reduces irrelevant background content and facilitates more effective feature extraction (see, e.g.,~\citep{crop2024}).

In our implementation, cropping is performed via contour-based thresholding. Each image is converted to grayscale, smoothed using a Gaussian blur, binarized with a low threshold, and subsequently refined through erosion and dilation operations. The bounding box of the largest connected component is then extracted, with a small pixel margin added for robustness. If the cropping procedure fails or results in an excessively small region, the original image is retained.

Following cropping, all images are resized to \(224 \times 224\) pixels to match the input requirements of ResNet-18. Channel-wise normalization is applied according to
\[
I_{\text{norm}} = \frac{p - 0.5}{0.5},
\]
where \(p\) denotes the raw pixel intensity.

A pretrained ResNet-18~\citep{He2016ResNet} network is employed to extract high-level feature representations from the MRI scans. The final fully connected layer of the network is replaced with an identity mapping, allowing it to output a 512-dimensional embedding vector:
\[
\bm{z} = f_{\theta}(I_{\text{norm}}), \qquad \bm{z} \in \mathbb{R}^{512},
\]
where \( f_{\theta} \) denotes the feature extraction function parameterized by weights \( \theta \). Feature extraction is performed in evaluation mode to prevent updates to the pretrained parameters.

All models operate directly on this standardized 512-dimensional feature space, ensuring that observed performance differences arise from architectural choices rather than variations in input dimensionality.

\subsection{Models and Evaluation Protocol}\label{app:models}
For comparison with the proposed QML approach, several classical machine learning models are considered, each operating on the same extracted feature vectors. All model hyperparameters are selected via an exhaustive grid search performed on the training data. The selected models are discussed as follows:

\paragraph{Linear Model (Logistic Regression).}
A Linear model utilizes a multinomial logistic regression classifier, which is a linear model. This model incorporates a single linear layer which directly maps the 512-dimensional input feature vector to the output class logits. A softmax function is used to calculate the probabilities. We used L2 regularization to prevent the overfitting. The model's parameters were optimized using the LBFGS solver, with a maximum of 500 iterations.
\paragraph{SVM (Support Vector Machine).}
Support Vector Machine (SVM) based on radial basis function (RBF) kernel used to identify non-linear relationships in the feature space. Non-linear decision boundaries are made possible by the kernel function's implicit mapping of input features into a higher-dimensional space. The kernel width is automatically scaled using the inverse of the feature variance ($\gamma = \text{scale}$), and the regularization parameter is set to $C=10$.
\paragraph{Random Forest (RF).}
Random Forest consisting of an ensemble of 300 decision trees. Each tree is trained using a bootstrap sample of the training data, and at each node split, a random subset of the input features is considered. This randomized training process lowers variance and promotes model diversity. The class probability outputs from each tree are averaged to determine the final prediction. To increase computational efficiency, the ensemble is trained in parallel.
\paragraph{MLP Baseline.}






A multilayer perceptron (MLP) baseline is employed, consisting of a shallow feedforward architecture with a single fully connected hidden layer of 512 neurons followed by a GELU activation function. The output layer is a linear fully connected layer mapping the hidden representation to class probabilities. This architecture is deliberately chosen to closely mirror the classical component of the hybrid quantum machine learning (QML) model. In this sense, the comparison between the MLP and the QML model can be interpreted as an ablation study: the quantum circuit in the hybrid architecture is removed and replaced by a purely classical hidden layer, while the remaining training protocol is kept unchanged. This allows us to isolate the contribution of the quantum layer and provides a fair and controlled comparison between the classical and quantum-enhanced approaches.

Both the MLP and the QML models are trained using the same optimization strategy and hyperparameters to ensure consistency. In particular, training is performed using the Adam optimizer with learning rate $\eta = 3 \times 10^{-3}$, batch size $32$, and $20$ training epochs. Categorical cross-entropy is adopted as the loss function:
\begin{equation}
\mathcal{L} = -\sum_{k=1}^C y_k^{(\mathrm{true})} \log p_k,
\end{equation}
where $p_k$ denotes the predicted probability of class $k$, obtained by applying the softmax function to the model outputs.

For both models, gradients are computed via standard backpropagation. Since the QML is trained on a classical simulator, gradient evaluation is carried out through automatic differentiation within the classical computational graph, without requiring parameter-shift techniques~\cite{mitarai2018quantum} (necessary in real hardware).


\paragraph{Validation strategies}

To validate the models, a 5-fold cross-validation procedure is employed. The full dataset $D$ is partitioned at each fold $k$ into a training set $D_{\mathrm{train}}^{(k)}$ and a validation set $D_{\mathrm{val}}^{(k)}$, following an $80\%/20\%$ split:
\begin{equation}
D = \bigcup_{k=1}^{5} \left( D_{\mathrm{train}}^{(k)}, D_{\mathrm{val}}^{(k)} \right).
\end{equation}

We first consider standard performance metrics. Classification accuracy is defined as
\begin{equation}
\mathrm{Acc} = \frac{1}{n} \sum_{i=1}^{n} \mathbf{1} \left( \hat{y}_i = y_i \right),
\end{equation}
where accuracy measures the proportion of MRI scans correctly assigned to their true tumor class. It provides a global indicator of classification correctness and is a commonly used baseline metric for multi-class medical image analysis.

The macro-averaged F$_1$ score is given by
\begin{equation}
F_1^{\mathrm{macro}} = \frac{1}{C} \sum_{c=1}^{C}
\frac{2 \cdot \mathrm{Precision}_c \cdot \mathrm{Recall}_c}
{\mathrm{Precision}_c + \mathrm{Recall}_c},
\end{equation}
which computes the harmonic mean of precision and recall for each class and then averages the result uniformly across classes. This metric is particularly important for medical datasets with class imbalance, as it prevents dominant classes from overshadowing minority tumor types.

In addition, we report the mean squared error (MSE) between the predicted class
probability vectors and the ground-truth labels:
\begin{equation}
\mathrm{MSE} = \frac{1}{n}\sum_{i=1}^{n}
\left\| \mathbf{p}_i - \mathbf{y}_{i}^{(\mathrm{true})} \right\|^2,
\end{equation}
which captures the quality and calibration of probabilistic
predictions, penalizing overconfident incorrect classifications even when the
predicted class label is correct.

We then apply the RGA, RGR and RGE metrics, and summarise them calculating the area under them, under different perturbations: data removal (AURGA), data perturbation (AURGR) and feature removal (AURGE).


The overall workflow proceeds as follows:

\begin{enumerate}
    \item Load MRI images from the dataset.
    \item Crop the images to remove non-informative background regions and focus on the brain area.
    \item Preprocess images by resizing to $224 \times 224$ pixels and applying normalization.
    \item Extract 512-dimensional deep features using a pretrained ResNet-18 model.
    
   \item Standardize the extracted 512-dimensional feature vectors.
\item Train multiple classifiers on the standardized features, including:
   \begin{itemize}
     \item Multi-Layer Perceptron (MLP),
     \item Random Forest (RF),
     \item Support Vector Machine (SVM),
     \item Linear Model,
     \item Quantum Machine Learning (QML).
   \end{itemize}
    \item Evaluate model performance using 5-fold cross-validation.
    \item Assess predictive performance using Accuracy and macro F1-score.
    \item Compute SAFE-AI reliability metrics—ranking accuracy (RGA), robustness to noise (RGR), and explainability robustness (RGE) along with their corresponding areas under the curves.
\end{enumerate}





\end{document}